# Toward human-centered AI-assisted terminology work

Antonio San Martín
Université du Québec à Trois-Rivières

Generative AI is likely to transform terminology work by creating new opportunities for automation. At the same time, it raises concerns about the future of terminologists and terminological resources, as efficiency pressures may encourage excessive automation based on the perception that human expertise can be replaced by AI. However, large language models remain unreliable for terminological purposes due to errors, hallucinations, and various forms of bias, making terminologists indispensable for ensuring the accuracy and reliability of terminological data. This paper argues that human-centered AI, an approach that emphasizes that AI's primary goal should be to contribute to human well-being, provides a framework for maximizing the benefits of generative AI while mitigating its risks. It contends that high levels of automation and meaningful human control are compatible and desirable, and that AI should enhance terminologists' capabilities while preserving their agency and decision-making authority. The implications of AI-assisted terminology work are examined through three interrelated dimensions: the augmented terminologist, ethical AI, and human-centered design. In particular, the paper addresses how AI integration reshapes the role of the terminologist, affects professional values and working conditions, requires the management of AI-generated bias, and calls for the design of AI tools around the terminologist's needs. The paper concludes that a human-centered orientation is necessary to ensure that AI strengthens, rather than undermines, the essential role of terminology work in supporting specialized communication and the accurate transmission of knowledge across languages and cultures.

## 1. Introduction

The rapid diffusion of generative artificial intelligence (GenAI) is transforming language-related professions. Translation has already been profoundly affected, prompting extensive scholarly debate. Terminology work (TW), by contrast, has historically engaged less directly with artificial intelligence (AI), despite its long-standing reliance on digital tools. This has changed with the emergence of large language models (LLMs), which can generate definitions, propose term equivalents, identify conceptual relations, and answer terminological queries in natural language. AI developments are therefore no longer peripheral to TW. However, the field still lacks principles to guide the integration of these technologies into professional terminological practice.

Terminological products[1] have an important societal role because they support specialized communication, translation, standardization, and knowledge transfer, with downstream linguistic, conceptual, social, and cultural effects. AI-mediated changes in TW thus influence what terms are prioritized and which knowledge perspectives are foregrounded or marginalized. While GenAI promises efficiency gains, unstructured adoption risks undermining the terminologist's professional expertise and autonomy, thereby contributing to bias amplification and the erosion of linguistic and conceptual diversity. Addressing this tension requires an explicit orientation that connects theoretical reflection with the concrete implications for TW.

This paper argues that human-centered AI (HCAI) provides a useful foundation for orienting AI adoption in TW. Existing research in translation studies has already drawn on HCAI to conceptualize the relationship between AI and professional practice. Drawing on this work and on broader AI scholarship, while accounting for the specific professional characteristics of TW, a human-centered framework for AI integration in TW is proposed. The objective is to provide a conceptual foundation for future research, inform the development of AI-assisted tools and workflows, and support the deployment of AI in TW in ways that preserve professional agency and the quality of terminological products.

The paper is organized as follows. Section 2 outlines the principles of HCAI. Section 3 reviews human-centered approaches to AI in translation studies. Section 4 develops a human-centered framework for AI-assisted TW organized around three complementary dimensions: the augmented terminologist, ethical AI, and human-centered design. Finally, the conclusion emphasizes the broader

1. Throughout the article, the term "terminological products" is used to refer not only to terminological resources (such as termbases and glossaries) but also to the results of ad hoc TW, which may include, for example, a single equivalent of a term in another language.

significance of a human-centered approach to AI in TW and identifies areas for future research.

## 2. Human-centered artificial intelligence

For decades, mainstream AI development has prioritized speed and efficiency over human values and societal impacts (Schmager et al. 2023, 1). This approach reflects a technology-centered vision that emphasizes machine performance rather than human purpose (Shneiderman 2020, 112). In response, HCAI calls for redefining the primary goal of AI as the promotion of human flourishing (Vallor 2024, 13).

Against this background, Shneiderman (2020, 112) metaphorically describes HCAI as a second Copernican revolution that restores humans to the center of the technological universe. AI is thus intended to revolve around human needs and values. This paradigm rejects the assumption that progress requires minimizing human involvement in favor of automation and instead defines AI's purpose as amplifying human agency and judgment, rather than replacing them. By redefining AI innovation as the enhancement of human capability instead of the displacement of human control, HCAI moves beyond "human-in-the-loop" configurations toward systems in which AI is in the loop: a supportive component that remains subordinate to human direction (Shneiderman 2020, 112).

This reorientation also requires acknowledging that AI does not operate in isolation. HCAI recognizes that intelligent systems function within sociotechnical ecosystems of "stakeholders, institutions, cultures, norms, and spaces" (Dignum and Dignum 2020, 2). Treating AI as a self-contained technological artifact overlooks the organizational and professional settings in which it is developed and used. As the Stanford Institute for Human-Centered AI (2025) emphasizes, technological progress must generate equitable impacts and broadly shared benefits. Responsibility therefore extends beyond developers to all actors involved in deployment (Dignum and Dignum 2020, 3), ensuring that control and accountability remain human.

Although HCAI encompasses multiple principles, two emphasized by Shneiderman (2020) are particularly important. The first holds that high automation and strong human control are compatible rather than opposed. Conventional AI thinking treats automation and control as a zero-sum trade-off, a position HCAI rejects. Shneiderman (2020, 115) illustrates this with smartphone cameras, which embody this synthesis of automation and human control. AI algorithms manage focus, aperture, and stabilization, while users retain control over framing, zoom, and timing. This example shows that automation can augment human capability without displacing human control, a core HCAI claim.

The second principle articulated by Shneiderman (2020) holds that AI should amplify rather than emulate human intelligence. Because artificial and human intelligence are qualitatively distinct, attempts at imitation misconstrue their respective natures. HCAI therefore frames AI as an instrument under human command that extends human capabilities. Progress is measured not by resemblance to human cognition but by the degree to which AI empowers human action.

HCAI has been studied across disciplines and perspectives. One prominent strand strongly associated with HCAI is ethical AI (Siau and Wang 2020), which holds that a human-centered approach must consider the rights and values of those who work with or are affected by AI systems (Capel and Brereton 2023, 9). From this perspective, HCAI entails designing AI systems that advance human well-being in alignment with human values, and incorporate safeguards against bias and discrimination. These concerns surface in concrete ethical challenges, including AI's environmental impact (Ding et al. 2025), intellectual property (Lucchi 2024), and the poor labor conditions underlying data annotation and other forms of invisible work sustaining AI development (Yılmaz and Bostancı 2025).

Another key component of HCAI is human-centered design (HCD), which constitutes its applied dimension. HCD places users at the center of AI development by prioritizing their needs and goals and involving them throughout the design process (Benyon 2019, 13). It also emphasizes human connection and the creation of inclusive solutions that account for diverse users and environments.

HCAI represents both a philosophical stance on the values that should guide AI development and deployment and an operational framework for designing AI systems that prioritize human agency and well-being. The following sections examine how these principles have begun to take shape within translation research and how they can, by analogy, inform AI integration into TW.

## 3. Human-centeredness in translation

Applied to translation, HCAI has given rise to two main lines of inquiry.[2] The first concerns the design of machine translation (MT) applications that address

2. Given the extensive body of literature on AI and translation, a comprehensive review falls outside the scope of this section. The discussion is thus limited to authors who have explicitly adopted a human-centered approach.

the needs of user groups beyond professional translators (Robertson et al. 2021; Carpuat et al. 2025). Because this paper concerns the application of HCAI to professional TW, attention is directed here to the second line of inquiry, which applies HCAI principles to the relationship between professional translators and AI-driven tools, including neural MT and, more recently, GenAI.

The application of HCAI to professional translation builds on earlier work on augmented translation (DePalma and Lommel 2017), which explores how technology can enhance translators' capabilities and offset inherent cognitive limitations (O'Brien 2024, 396). Under the right circumstances, augmentation can enable translators to work more efficiently and often produce higher-quality output than without technological support. The range of technologies contributing to this augmented environment includes the internet, word processors, translation memories, MT, quality estimation systems, automated content enrichment, terminology management tools, project management, translation management systems, corpus analysis tools, bitext aligners, and GenAI (Lommel and DePalma 2021; Bernardini 2022; O'Brien 2024; Rivas Ginel and Moorkens 2024).

However, in some contexts, certain technological tools are imposed on translators (Jiménez-Crespo 2025a, 3) and have been used in the industry primarily to increase speed and efficiency while reducing costs (Lommel and DePalma 2021). Under what Moorkens (2020) terms "digital Taylorism," translators become "cogs in a large machine," as automation is maximized and work is fragmented into small units. As a result, technological augmentation has often advanced at the expense of translators' needs and well-being (O'Brien 2024, 399). Human-centered approaches to translation seek to redress this imbalance.

Within an HCAI framework, O'Brien (2024, 392) argues that the human–machine dichotomy in translation is no longer meaningful, as both function as interdependent elements within a shared cognitive and technological ecosystem. She thus proposes replacing antagonism with a notion of kinship, emphasizing symbiotic collaboration rather than competition (393). This shift was anticipated in her earlier work, which framed translation as a form of human–computer interaction and highlighted the central role of digital tools in contemporary professional practice (O'Brien 2012).

Drawing on Shneiderman (2020), O'Brien (2024, 393) argues that AI should augment translators' abilities while leaving full control with humans. She further contends that high automation can coexist with strong human control in translation and that this configuration is both feasible and desirable (394). As Jiménez-Crespo (2024, 279) observes, this position is especially pertinent in light of translators' reports of diminished agency and fears of replacement by AI (Cadwell et al. 2018; Moorkens 2020).

Building on Shneiderman's (2020, 116) distinction between emulation and empowerment, O'Brien (2024, 394) adds that AI translation technologies should prioritize empowering translators rather than accelerating the emulation of human translation at the expense of user experience. She further maintains that translators' documented resistance (Cadwell et al. 2018), ambivalence (Koskinen and Ruokonen 2017), and disillusionment (O'Brien et al. 2017) toward translation technology stem from this focus on speedier emulation instead of empowerment.

Jiménez-Crespo (2024, 279) further attributes this resistance to translators' limited involvement in the design and implementation of computer-assisted translation tools. Briva-Iglesias (2024, 164) notes that technology adoption in translation has often proceeded by first developing tools and then training translators to adapt to them. From this perspective, Jiménez-Crespo (2024, 279) argues that an HCAI approach is essential because it recenters translators in technological development and decision-making. This requires integrating translators' perspectives into the design of AI tools and workflows so that technologies align with professional practice rather than being imposed on it, thereby empowering translators instead of diminishing or replacing them.

As Jiménez-Crespo (2025b, 8) highlights, this human-centered orientation aligns with the recognition of the human factor as a core component of contemporary translation studies. It also resonates with Hu's (2004) distinction, later popularized by Chesterman (2009), between translator studies, which focus on the human dimensions of the translation process, and translation studies, which center on the translated product.

Jiménez-Crespo (2024, 2025a) has contributed empirically to advancing HCAI in translation by examining translators' perceptions of control and autonomy. Based on a survey of 50 translators working in the USA, he shows that respondents anticipate a decline in their ability to control the integration of AI into their work. This perceived loss of agency is attributed not to AI itself but to its top-down imposition by external stakeholders such as language service providers and clients. Translators therefore express a strong desire to retain decision-making authority over the technologies used throughout the translation process. Given their ultimate responsibility for the final product, they also want the ability to tailor tool behavior to their needs. At the same time, they report that their articulated needs are often not considered by other actors in the translation ecosystem.

Beyond control and autonomy, translation research has examined risk from a human-centered perspective. Using medical translation as a high-stakes case, Tercedor-Sánchez (2025) demonstrates the continued indispensability of human translators. She shows that AI-generated translations entail significant risks when machines fail to adapt lexical choices to the communicative situation and to the needs of users. Her analysis highlights the risk of reproducing bias in the absence of professional mediation and underscores the importance of managing terminological variation and metaphorical language, areas in which the current capabilities of AI systems remain limited.

In sum, applying the HCAI paradigm to translation calls for reconceptualizing human–machine relations as symbiotic, with AI serving a subordinate role to the human. Despite persistent fears of replacement, human translators remain indispensable, as translation relies on forms of judgment and expertise that cannot be delegated to machines. As Jiménez-Crespo (2025b, 13) argues, HCAI redirects attention to the fundamental human capacities that cannot be replaced but only augmented. These insights raise broader questions about how human-centered principles should guide AI adoption across adjacent language professions. The following section extends this perspective to TW.

## 4. Human-centered AI-assisted terminology work[3]

Compared with TW, translation has engaged with AI for a longer period, particularly since the advent of neural MT in the mid-2010s (Stahlberg 2020). This earlier exposure has necessitated sustained consideration of AI's impact on professional practice. Because TW has had comparatively limited exposure to AI, the advent of GenAI has created an urgent demand for a framework to guide its integration and orient research at the intersection of terminology and AI. The HCAI paradigm (Shneiderman 2020; Capel and Brereton 2023; Schmager et al. 2023) is well positioned to fulfill this role.

GenAI may become one of the most significant technological developments in TW since the advent of termbases. While it promises gains in efficiency and quality, it also introduces new risks and intensifies concerns that AI could replace terminologists or even supplant terminological products.

For many users seeking terminological information, querying an LLM-based chatbot such as ChatGPT or Claude offers practical advantages over consulting traditional terminological resources. Users can provide textual or graphical context, receive responses tailored to their needs, and ask follow-up questions for clarification or additional detail. Through retrieval-augmented generation (RAG), LLMs can also provide up-to-date information from the web or other sources. Moreover, they eliminate the preliminary step of identifying an appropriate terminological resource, as they can address virtually any type of query.

3. This section focuses exclusively on LLMs due to their growing integration into terminological workflows. Other machine learning applications are excluded, as such tools have seen limited adoption in TW, primarily because most require coding skills and lack user-friendly interfaces.

However, relying on LLM-based chatbots for terminological information has significant limitations that render human terminologists indispensable. Chief among these are errors and hallucinations, which undermine trust, as well as the potential reproduction of bias. In this context, terminologists and terminological products offer an important advantage: reliability grounded in expert human judgment. Validation by terminologists produces authoritative content that machines cannot replicate. In commercial and institutional settings, termbases contain approved terminology tailored to specific organizational environments, whereas in standardization and terminology planning, resources reflect negotiated human consensus. In all cases, LLM-based chatbots cannot match the accuracy and expertise that underpin professionally developed terminological products.

Paradoxically, terminological data validated by human experts are also valuable for LLMs. As Khemakhem et al. (2025) note, designations, definitions, contexts, and related data categories provide structured, high-quality knowledge that supports AI-mediated specialized communication. Such information can be integrated into language models through training, fine-tuning, prompting, or retrieval, thereby improving terminological precision, strengthening factual grounding, and reducing hallucinations. An emerging development in this area is terminology-augmented generation (Di Nunzio 2025; Lackner et al. 2025), which incorporates domain-specific terminologies and associated data into LLM outputs via retrieval mechanisms analogous to RAG.

From this perspective, GenAI is relevant to TW primarily because it can assist terminologists in two modalities: post-editing TW[4] and intelligent support (San Martín 2024). In practice, however, the boundary between these two categories is fluid, as many workflows combine elements of both.

Post-editing TW refers to the approach in which terminologists refine and validate draft terminological content generated by the machine. A paradigmatic case is definition writing, which until the advent of GenAI had resisted usable automation. AI systems can now produce initial definitions, typically guided by carefully crafted prompts, which terminologists subsequently refine before dissemination. This post-editing logic also extends to other types of terminological data, including conceptual relations, term equivalents, denominative variants, and usage examples.

By contrast, intelligent support encompasses tasks in which AI assists the terminologist without generating content intended for post-editing and dissemination. This category includes:

4. The term “post-editing terminology work” is based on “post-editing lexicography” (Jakubíček et al. 2018).

- helping terminologists become familiar with a given domain;
- supporting the interpretation of specialized texts and the clarification of unfamiliar concepts;
- performing term extraction;
- analyzing existing definitions to identify semantic traits that can inform definition writing and conceptual modeling;
- assisting corpus analysis, for example, by examining concordance lines to extract linguistic or conceptual information;
- supporting term creation;
- enhancing quality assurance by reviewing the terminologist's work, flagging potential errors, and suggesting improvements;
- assisting terminologists in coding scripts that help automate tasks, notably with agentic systems such as OpenAI Codex and Claude Code.

The potential of LLMs to support TW has been examined by numerous authors. Most studies are small-scale and focus on evaluating model performance across specific terminological tasks. Term extraction is the most extensively examined capability. Existing studies suggest that, despite imperfections, LLMs offer promising performance for this task (Giguere and Iankovskaia 2023; Łukasik 2023; Bezobrazova et al. 2024; Di Nunzio et al. 2024; Heinisch 2025a; Liu et al. 2025; De Pourcq et al. 2025; Tran et al. 2025). Other terminological tasks have also received attention. Research on definition generation (Bezobrazova et al. 2024; Di Nunzio et al. 2024; Nahod and Vukša Nahod 2024; Varga 2024; Heinisch 2025a) and definition extraction (Heinisch 2025a; Tran et al. 2023) has identified several limitations in LLM performance. However, the small number of studies examining these tasks, together with their generally limited scale, precludes broader generalizations about the capabilities of LLMs in this area.

Another aspect of LLM performance that has been evaluated is their capacity to propose key terms within specific domains (Heinisch 2025a; Vidal Sabanés and Da Cunha 2025). While Vidal Sabanés and Da Cunha report high accuracy for ChatGPT-4 in Spanish, Heinisch finds that several LLMs confuse regional variants and even invent terms in German. Vidal Sabanés and Da Cunha also assess ChatGPT's ability to propose synonyms and select the clearest option; although promising, its performance falls short of the manually constructed gold standard. Speranza (2025) evaluates the identification of term relations in English by ChatGPT and Gemini and shows that first-level conceptual relations are handled adequately, while performance declines sharply at deeper relational levels. Finally, Nahod and Vukša Nahod (2024) test ChatGPT-4's ability to generate term equivalents between Croatian and English and report a success rate of only about 70% in both directions.

Taken together, these studies show that LLMs do not yet deliver consistently reliable results, making human intervention essential. Their imperfections, however, do not negate their usefulness. Other tools routinely used by terminologists, such as hybrid term extractors or word sketches in Sketch Engine, are likewise imperfect, yet remain integral to professional practice. Overall, the evidence indicates that LLM performance varies considerably depending on the task being evaluated, the languages and domains concerned, and the model employed. LLM output is therefore promising but uneven in its accuracy.

Contemporary professional environments increasingly prioritize productivity and cost efficiency. When combined with the unfounded belief that AI can fully replace terminologists, these pressures risk devaluing or displacing the terminologist's expertise. A human-centered approach to AI integration is therefore not merely desirable but also necessary for TW. The remainder of this section examines how HCAI can be applied to TW along three main axes: the augmented terminologist, ethical AI, and human-centered design.

## 4.1 The augmented terminologist

Like translators, terminologists have benefited from technological augmentation for decades, particularly since the emergence of electronic databases in the second half of the twentieth century (Schmitz 2005, 3). Beyond terminology management systems, a range of digital tools amplifies terminologists' capabilities. These include the internet, corpus analysis tools, monolingual and bilingual automatic term extractors, and bitext aligners. Together, these tools have expanded what individual terminologists can achieve in both volume and precision.

TW is thus inseparable from technology; the contemporary terminologist is therefore an augmented terminologist. The advent of GenAI, however, has prompted a reassessment of automation, as AI can be integrated into many stages of the terminological process. Tasks once limited to manual execution, such as drafting definitions or analyzing concordances, can now be automated with varying success. Moreover, AI systems can generate entire glossaries or populate terminological database entries, albeit with uneven accuracy, raising concerns about the future role of the terminologist.

In this context, Shneiderman's (2020) ideas on the compatibility of high automation with strong human control and the emphasis on empowerment over emulation are significant for TW. The following subsections explain how these two principles apply to TW before discussing the new skill set that terminologists will need in order to work effectively within AI-enhanced workflows.

### 4.1.1 *High levels of automation with strong human control*

High automation and strong human control in terminological workflows are not only compatible but also desirable. A terminologist supported by well-designed AI may outperform both standalone AI output and purely manual work, as human expertise and machine efficiency operate synergistically (Capel and Brereton 2023, 6). This is particularly important in the current global context, where new concepts and terms emerge at unprecedented speed (Shen et al. 2021, 48), and terminological resources must keep pace (Archibald et al. 2020, 5). However, automation entails risks (discussed in the following sections) that can be mitigated only if terminologists exercise active oversight and decision-making authority, thereby positioning themselves at the center of an "AI-in-the-loop" model.

For AI to enhance TW, an appropriate balance between automation and human control is required. Both extremes are counterproductive: too much automation risks degrading the quality of the final product, while too much human intervention can slow work unnecessarily (Shneiderman 2020, 116). This balance must be established at the subtask level. For example, in definition generation, the terminologist maintains control by drafting the prompt that guides the process, specifying elements such as the target audience and the perspective to adopt (i.e., the contextual constraints to apply to the definition (San Martín 2022)). The terminologist may also supply source texts for the LLM to use via RAG. Automation is then applied, with the LLM generating a draft definition. Subsequently, the terminologist resumes control to manually review and validate the output. At this stage, automation can again assist by helping the terminologist resolve uncertainties or by streamlining corpus analysis to support the validation process.

In practice, decisions about how to allocate automation and human intervention are highly context dependent. They may vary depending on the setting (commercial, institutional, academic, etc.), the objectives (description, standardization, translation, terminology planning, etc.), the specific project, the languages involved, and even the individual preferences of the terminologist. It is equally important to assess which AI or non-AI tools are best suited for each task or subtask, as LLMs are not the only tools available to terminologists. Non-AI tools, such as word sketches in Sketch Engine for collocation analysis (Kilgarriff and Tugwell 2001) or hybrid term extractors (Drouin 2003), continue to play a valuable role in the terminological workflow.

Establishing this balance requires systematic performance-oriented research. This includes benchmarking LLMs across various tasks and subtasks of TW to identify where they perform well and where they fall short,[5] bearing in mind that performance can vary depending on domain and language. These evaluations must be sufficiently complex to reveal how AI handles intricate semantic phenomena in terminology, such as denominative variation (Freixa 2006), contextual variation (San Martín 2022), multidimensionality (Bowker 2022), polysemy (Vezzani et al. 2025), and metaphor (Tercedor-Sánchez et al. 2012). Benchmarking is also valuable for providing further evidence that terminologists remain indispensable by identifying the uniquely human capabilities that cannot be delegated to AI. Such findings will help terminologists claim their irreplaceable role in this evolving technological landscape.

Research should also actively seek input from terminologists about the types of augmentation they find valuable and their preferred forms of implementation (O'Brien 2024, 400). For example, they may favor the integration of LLMs into corpus analysis tools (where they can always return to specific concordance lines) over the fully automated population of fields in a terminology management system.

Attention should also be given to how AI augments TW within situated practice. This involves researching how terminologists interact with AI tools across different phases of their work and assessing whether these interactions truly amplify their capabilities. The focus should be not only on efficiency and quality, but also on whether AI contributes to the empowerment of terminologists, as discussed in the following section.

#### 4.1.2 *Emulation vs. empowerment*

According to Shneiderman (2020, 116), a long-standing goal in AI research has been to emulate human intelligence. While emulation can be helpful in specific cases, an HCAI perspective holds that the primary objective of AI development should be to empower humans, and that emulation should be pursued only when it supports that aim. When applied to TW, this implies that AI systems should prioritize augmenting terminologists' capabilities while ensuring meaningful human control and a high-quality user experience (see Section 4.3).

AI and human intelligence are inherently different, and each has its own strengths and capabilities (Ozmen Garibay et al. 2023, 392). AI excels at processing large amounts of data and identifying patterns across multiple tasks with remarkable speed and memory capacity. Human intelligence, by contrast, involves embodied cognition, situated reasoning, real-world perception, and creative

5. Excelling and failing is not a black-and-white matter, as LLMs may provide correct answers inconsistently or only to a certain extent.

insight, none of which can be reduced to data patterns (Robbins and Aydede 2012; Shapiro 2019; Lockhart 2025).

Proposing candidate term equivalents is a clear example of a terminological task in which an LLM can exploit its distinctive strengths. By drawing on its training data and powerful pattern recognition abilities, an LLM can suggest plausible candidates across languages with remarkable speed, sometimes proposing term equivalents that would be time-consuming or even highly challenging for a human to identify manually. This is valuable in TW, since verifying the adequacy of a candidate equivalent is usually much faster than discovering it from scratch.

After identifying a potential equivalent, the system can emulate one aspect of human practice by checking whether the proposed term is attested. In this case, an AI system can search the web or query a corpus provided by the terminologist via RAG to gather real-world usage evidence and present the results for review. However, this process does not fully replicate a human action. RAG leverages computational capabilities that humans lack, such as fast information extraction from large amounts of text.

Regardless of the nature of the automated task, the central goal is the empowerment of the terminologist, which entails both amplifying their capabilities and preserving their agency. To this end, terminologists should be able to choose to verify independently if the term is attested or to delegate this step to the system. They should also have the possibility of adjusting the parameters that guide the RAG process, such as indicating which websites to consult or which corpus to use. Ultimately, the terminologist must retain full authority over the final decision, since they are responsible for determining whether the term proposed by the LLM is conceptually equivalent to the source term.

### 4.1.3 *AI literacy and new skills*

The integration of AI into terminological workflows is expected to transform the skill set required for TW, reshaping what makes a terminologist a valuable professional. As AI takes on a growing role, terminologists will have to acquire new competencies, while some traditional skills may become less relevant or less frequently used.

Many of these emerging competencies can be grouped under the umbrella of AI literacy, defined as "a set of competencies that enables individuals to critically evaluate AI technologies; communicate and collaborate effectively with AI; and use AI as a tool online, at home, and in the workplace" (Long and Magerko 2020, 2).

Drawing on the AI literacy skills outlined by Annapureddy et al. (2025) and Zhang and Magerko (2025), several emerging competencies that terminologists will need to develop can be identified:

- understanding how different AI models can support terminological tasks, while keeping pace with rapidly evolving capabilities and limitations;
- recognizing the strengths and weaknesses of AI tools to judge when their use is appropriate, selecting the most suitable tool, and determining the appropriate automation level for each task or subtask;
- using a broad range of AI-enabled tools with efficiency and confidence;
- formulating prompts that allow LLMs to generate the desired output;
- understanding the ethical and legal implications of AI use;
- evaluating AI-generated content critically, anticipating that errors or biases may arise and must be detected and corrected (see Section 4.2.2).

By analogy with Jiménez-Crespo's (2025c, 282) argument about translators, terminologists will likewise need to cultivate the capacity to explain to stakeholders why their expertise remains indispensable, even in highly automated environments. Terminologists have long had to actively claim the importance of their work across diverse professional settings (Drame 2015, 508; Warburton 2021, xxi), yet the rise of AI heightens this challenge.

At the same time, the acquisition of new AI-related skills should not come at the expense of the core components of a terminologist's professional competence, as Jiménez-Crespo (2025c, 279) has highlighted for translators. This underscores the importance of preserving meaningful human control throughout AI-enhanced terminological workflows, since reducing the terminologist's role to that of a simple validator of AI output would ultimately risk professional deskilling.

## 4.2 Ethical AI

As previously noted, ethical AI takes into account the rights and values of all individuals working with or affected by AI systems (Capel and Brereton 2023, 9). In the context of TW, this includes the terminologist as well as a broader range of individuals who may be affected, either directly or indirectly, by terminological decisions made with AI involvement. This group includes users of terminological products as well as society at large. Although ethical AI covers a wide range of issues, two dimensions are particularly salient for TW: the protection of the terminologist's values and well-being, and the mitigation of bias.

### 4.2.1 *The terminologist's values*

AI adoption never occurs in isolation but is always embedded in a sociotechnical context that encompasses institutional, cultural, and normative dimensions (Dignum and Dignum 2020, 2). This context brings together a diverse set of stake-holders, whose configuration varies across settings. They may include entities such as companies, language service providers, standardization bodies, academic institutions, and publishers, as well as individuals such as terminologists, project managers, translators, technical writers, and all end users. These configurations shape how terminologists interact with AI systems and how such systems influence TW. Under traditional approaches to AI, machine autonomy and algorithmic performance have often been treated as primary objectives (Shneiderman 2022, 7). While these goals are not inherently problematic, they become ethically questionable when pursued at the expense of the terminologist's values and working conditions. The amplification of terminologists' capabilities through AI cannot be considered human-centered if it prioritizes economic considerations over the social benefits of TW.

Human-centered research foregrounds the terminologist's experience and situated context, examining how terminologists engage with AI technologies in practice (Capel and Brereton 2023, 13).[6] In this respect, just as translator studies have emerged as a branch within translation studies (Chesterman 2009), terminology studies would benefit from a corresponding branch that might be termed *terminologist studies*. Such an approach calls, among other things, for the empirical identification of terminologists' values and professional priorities. While this research remains to be conducted, it is nevertheless possible to hypothesize a set of core values that underlie terminological practice beyond professional autonomy:

- linguistic and conceptual precision;
- preservation of linguistic and epistemic diversity through which specialized knowledge is structured;
- fair and accurate representation of the worldviews that shape meaning in specialized domains;
- facilitation of specialized knowledge communication across languages and cultures;
- promotion of universal access to up-to-date expert knowledge through unbiased terminological representations that accommodate different levels of specialization.

For terminologists to uphold their values, the sociotechnical context must be configured to enable AI systems to be developed and integrated into terminological workflows while preserving terminologists' agency. This requires the active involvement of terminologists in the design and implementation of the tools they

**6.** This line of research has already been initiated by Wissik (2025), who studied how terminologists in institutional settings are beginning to integrate GenAI into their workflows.

are expected to use (see Section 4.3). As translators have argued in their own field (Jiménez-Crespo 2025a, 13), terminologists likewise should be able to make all technological decisions and retain authority over the final product. Top-down impositions of AI systems or workflows are therefore incompatible with human-centeredness.

When AI is introduced primarily to maximize productivity or reduce costs without regard for the needs and values of terminologists, the resulting autonomy loss can lead to demotivation and increased stress (Oviatt 2021, 278).[7] In such cases, AI integration is unethical because it undermines the psychological and occupational well-being of terminologists and, more broadly, because the failure to preserve the terminologist's values can compromise the societal contribution of terminological products as instruments of specialized communication and knowledge transfer. The following section examines how terminologists play a key role in ensuring the ethical quality of terminological products through the management of AI bias.

#### 4.2.2 *Bias*

While bias in LLMs can arise from various sources, three key factors are particularly relevant (Ferrara 2023): the composition of the training data, the functioning of the underlying algorithms, and the post-training phase that shapes the model's behavior.

The introduction of bias in LLMs begins with the selection of training data. Because LLMs require massive textual corpora, developers primarily rely on web data, which are largely uncurated and only minimally filtered to remove low-quality content (Ferrara 2023). The precise composition of the training data used for major proprietary LLMs, such as those behind ChatGPT or Claude, remains undisclosed (Ravichander et al. 2025, 1962). Nevertheless, available evidence suggests that the Common Crawl dataset constitutes the largest component (Baack 2024). In its CC-MAIN-2025-47 release, Common Crawl contained 42.08% English-language content, followed by Russian at only 6.48% (Common Crawl 2025a). It is also estimated to be dominated by US-based content.[8]

7. This has been demonstrated in the case of translators by Sakamoto et al. (2024).

8. There are no data on the geographic origin of the texts in Common Crawl, but a filtered version known as the Colossal Clean Crawled Corpus is estimated to consist of 51.3% of content hosted in the United States, followed distantly by India at 3.4% (Dodge et al. 2021, 1289). Although this does not directly indicate the geographic origin of the texts, it suggests that a large proportion of the training data used by major proprietary LLMs is not only in English but also predominantly of US origin.

Unlike terminologist-built corpora, which ideally follow strict quality and representativeness criteria (Bowker and Pearson 2002, 45), proprietary LLM training data are unbalanced in terms of text genres and domains (Navigli et al. 2023, 3). This genre imbalance becomes apparent in the three most represented web domains in Common Crawl: two blog platforms (Blogspot and WordPress) and Wikipedia (Common Crawl 2025b). As an illustration of the domain imbalance, Navigli et al. (2023, 4) note that the most represented domains in the English Wikipedia by article count are sports, music, and geographic locations, which substantially outnumber domains such as physics, mathematics, and chemistry.

A second source of bias arises from algorithmic functioning, which can amplify distortions in the training data. Because LLMs replicate statistical patterns found in their input data and are optimized to generate high-probability outputs, they tend to reinforce and overrepresent dominant patterns (Sourati et al. 2026, 5). For instance, since English texts dominate the training data, the cultural knowledge embedded in those texts is likely to appear in LLM outputs in an even higher proportion than in the original data.

Post-training processes such as reinforcement learning from human feedback (RLHF) further shape model bias. In RLHF, human reviewers evaluate outputs and steer the model toward responses deemed appropriate or helpful. Although intended to mitigate harmful biases (McIntosh et al. 2024, 1572), these evaluations reflect the sociocultural context in which they are embedded. Because these processes are often grounded in North American norms, post-training can further skew LLM output toward US-centered worldviews (Ryan et al. 2024, 16122).

The factors mentioned above give rise to different types of bias, which are discussed separately below, although they often overlap in practice. If these biases are not mitigated by terminologists, they can lead to a range of detrimental consequences.

*Linguistic bias*

Linguistic bias refers to the undesirable effects of LLMs favoring certain languages or linguistic features. Since training data are predominantly in English, especially US English, AI-generated terminological content is expected to be more accurate in this language. A subtler effect of this bias is cross-linguistic interference, whereby a language (generally English) influences outputs in other languages. In fact, studies suggest that LLMs may rely on English as a pivot language (Wendler et al. 2024; Y. Zhao et al. 2024), that is, as an intermediate language through which information is processed when generating content in another language. Rigouts Terryn and de Lhoneux (2024) report that, in their experimental setup, 16% of linguistic errors in French and Dutch texts generated by GPT-4 and two smaller LLMs show clear English influence.

In AI-assisted TW, this bias implies that, for languages other than English, LLMs may invent terms through literal translation from English or privilege English-motivated denomination patterns over established target-language conventions. English interference may also affect definitions and usage examples, producing incorrect terms as well as grammatical and stylistic errors.

From a conceptual perspective, linguistic bias may lead LLMs to project Anglo-American conceptual systems onto other languages, disregarding anisomorphism, whereby languages structure specialized domains differently (León-Araúz 2022, 478). This effect can appear in any LLM-generated terminological output, including term equivalents, definitions, contexts, and conceptual relations.

Linguistic bias can also be intralingual, occurring when terms and conceptual systems from dominant regional variants are projected onto less-resourced ones. For example, as Heinisch (2025a, 17) noted, when three different LLMs (including GPT-4o-mini) were prompted to suggest Austrian terms, they produced some terms originating from Germany.

Finally, Tercedor-Sánchez (2025, 98) identifies another form of linguistic bias: because LLMs favor dominant patterns, they underrepresent low-frequency denominative variants in their suggestions. Yet such variants may be motivated by dialectal, functional, discursive, interlinguistic, and cognitive factors (Freixa 2006, 69), which may make these terms more appropriate in specific contexts.

*Temporal bias*

Temporal bias arises when LLMs fail to adequately represent the time-related dimension of specialized knowledge. Two factors contribute to this problem. First, training data often combine texts produced at different points in time without clear temporal distinctions (B. Zhao et al. 2024, 15015). Second, training-data cutoffs prevent models from incorporating the most recent developments. Consequently, temporal bias may manifest in the presentation of diachronic data as synchronic, or vice versa, as well as in the failure to reflect recent conceptual changes and terminological innovations. This is particularly problematic in TW, as specialized knowledge evolves rapidly, and terminological representations must accurately reflect the current state of a domain at both the conceptual and linguistic levels.

*Cultural bias*

Culture influences language, including terminology. Specialized language encodes cultural knowledge (León-Araúz and Faber 2024, 39), and terms are conceptualized differently depending on the cultural context. Cultural bias is therefore a central concern in AI-assisted TW.

Current proprietary LLMs are trained on datasets in which Anglo-American and, more broadly, Global North cultures are disproportionately represented. As a result, the cultural knowledge embedded in these contexts is more likely to be reflected in model outputs than are alternative cultural perspectives. The tendency of LLMs to reproduce such cultural distortions has already been documented by Tao et al. (2024). Consequently, these models may privilege certain culturally motivated denominative variants while predominantly reflecting Global North cultural knowledge within definitions and concept systems.

In AI-assisted TW, this bias can lead to terminological representations that obscure culturally grounded conceptual distinctions or marginalize locally situated knowledge, notably in domains where expertise and practice are particularly embedded in specific cultural and geographical contexts.

*Epistemic bias*

Epistemic bias refers to distortions in the representation and organization of knowledge. Current proprietary LLMs overrepresent mainstream knowledge by amplifying what is most prevalent in their training data. LLM output can be interpreted as knowledge structured around prototype-like representations in the sense of Cognitive Linguistics (Rosch 1978), a feature that may initially appear desirable. However, prototypes are context-dependent rather than universal (San Martín 2022, 2), and LLMs are structurally skewed toward a specific range of knowledge.

LLMs privilege dominant scientific paradigms tied to Global North institutions and prevailing schools of thought, as well as knowledge situated within well-documented domains. Consequently, they underrepresent or invisibilize alternative epistemologies, Global South knowledge production, emerging fields, and diverse conceptual frameworks.

Epistemic bias has direct implications for terminological decision-making, as term choices, definitions, contexts, and conceptual relations generated or supported by LLMs may reproduce dominant epistemic hierarchies rather than reflect the full diversity and dynamicity of specialized knowledge.

*Social bias*

Social bias involves prejudicial views, stereotypical representations, and discriminatory attitudes arising from historical and structural power asymmetries and directed toward specific social groups based on factors such as gender, race, ethnicity, age, sexual orientation, and disability (Gallegos et al. 2024, 1098; Navigli et al. 2023, 7).

In the context of TW, such bias can appear insidiously in AI-generated suggestions for terms, definitions, usage examples, and conceptual relations. If the resulting terminological products are not debiased, they can have a direct social impact by perpetuating negative or distorted perceptions of disadvantaged individuals (Tercedor-Sánchez 2025, 95). From a terminological standpoint, this would compromise the ethical responsibility of terminologists to produce accurate and socially responsible terminological resources.

*Bias mitigation*

The different forms of bias discussed above have distinct consequences. Managing linguistic and temporal bias primarily involves ensuring the accurate representation of specialized language and the knowledge it conveys. In contrast, addressing cultural and epistemic biases calls for the fair representation of specialized knowledge and sensitivity to the diversity of perspectives through which it is structured. Addressing social bias additionally requires avoiding stereotypical or discriminatory depictions of social groups. If disseminated without validation by a terminologist, AI-generated terminological data risk reinforcing these biases. LLMs are especially prone to inaccuracies and hallucinations when generating terminological content in languages other than English or when dealing with non-dominant cultures and conceptual frameworks. As a result, both the quality of AI-generated suggestions intended for post-editing by terminologists and the reliability of LLMs in providing intelligent support for terminological decision-making may be compromised.

If terminologists do not retain full decision-making power over the integration of AI in all aspects of terminology management and dissemination, inaccurate, hallucinated, contextually inappropriate, or biased terminology (as well as the specialized knowledge conveyed by terms) may circulate and be treated as authoritative. These terms may then be used in real communicative acts, such as specialized texts and translations. The issue may be further amplified when such texts inform other terminological products, for example, by entering corpora used by terminologists, or when those texts are included in the training data of LLMs. This may create a feedback loop that accelerates linguistic and conceptual distortion.

AI implementation raises ethical concerns if it amplifies or perpetuates any form of bias. In AI-assisted workflows, terminologists can mitigate bias both upstream and downstream. Upstream, this involves using effective prompting strategies and constraining RAG to contextually adequate sources. Downstream, it requires critically reviewing AI outputs using expert judgment, supported by the analysis of representative corpora, the consultation of diverse sources, and even the use of an LLM to assist in the debiasing process.

As Tercedor-Sánchez (2025, 95) notes, debiasing is challenging for both machines and humans. Research in terminology can contribute to bias mitigation by evaluating LLMs with respect to different types of bias in terminological tasks, thereby enabling terminologists to better anticipate systematic distortions and exercise more effective critical control over AI-generated terminological outputs.

## 4.3 Human-centered design

The integration of GenAI into terminological tools remains marginal at present, leaving terminologists to interact with GenAI mainly through standalone chatbots. This mode of access creates unnecessary friction, including interface switching and repeated copying and pasting. From a human-centered perspective, AI functionalities should instead be embedded directly within terminological tools. Alternatively, agentic AI environments could function as a unified interface through which terminologists can operate specialized tools.[9] Crucially, such integrations are not merely a technical issue but a design challenge that should follow established principles in human–computer interaction (HCI).

Given that TW is inseparable from the use of digital tools, it already constitutes a form of HCI, similar to translation (O'Brien 2012). Within HCI, human-centered design (HCD) (Giacomin 2014) is a core component of HCAI (Pyae 2025, 9) that addresses the cognitive, emotional, social, and organizational factors of human–machine interaction (Boy 2013, 197; Giacomin 2014, 612; Lahlou 2017, 167). Applied to AI-assisted TW, HCD principles (Benyon 2019, 13) imply that the design of AI applications and their integration into terminological workflows should revolve around four principles:

1. *Prioritization of the terminologist's needs.* AI systems and interfaces should be designed around the terminologist's "needs, requirements, expectations, and desires" (Tosi 2020, 4). This principle seeks to prevent reverse adaptation (Winner 2020, 174), in which systems are developed independently of user needs, and professionals are then forced to adjust their practices to the system rather than the system being adapted to their tasks and constraints. As Jiménez-Crespo (2024, 279) notes for translators, alignment with professional needs reduces resistance and enables practitioners to fully leverage AI advances.
2. *Involvement of terminologists in the design process.* Terminologists should be actively involved in the conception, design, integration, and refinement of AI

9. Agentic systems such as Codex and Claude Code can function as a coordinating interface through which terminologists can access and use different tools and services through mechanisms such as the Model Context Protocol (MCP). By combining conversational interaction with coding capabilities, these platforms can orchestrate external resources, including corpus analysis tools and terminological databases, process and transform terminological data, and automate complex workflows.

tools (Vallor 2024, 17), in line with participatory design principles (Bødker et al. 2022). This involvement is essential because terminologists are both the primary users of these systems and the domain experts best placed to assess their suitability for terminological workflows (Capel and Brereton 2023, 12). Systematic collection of terminologist feedback is also necessary to refine AI systems over time and maintain alignment with evolving needs (Ozmen Garibay et al. 2023, 392).

3. *Communication and collaboration*. TW is inherently shaped by communication and cooperation among stakeholders. AI applications should, therefore, be designed to support interaction among terminologists and facilitate coordination with other relevant actors, such as managers, domain specialists, translators, and technical writers.
4. *Design for diversity*. AI design must account for the diversity of contexts in which TW takes place, including variation across domains, languages, cultures, and organizational settings. A human-centered approach also requires explicit attention to accessibility and inclusivity (Benyon 2019, 105).

As discussed in previous sections, a central concern in human-centered AI-assisted TW is the preservation of human control. In this respect, insights can be drawn from Jiménez-Crespo's (2024, 2025a) work on translators' perceptions of control and autonomy. If his findings are transposed to terminological practice, it can be hypothesized that terminologists will experience a stronger sense of control when tools combine robust usability with mechanisms that allow them to selectively activate or deactivate specific system functions (Jiménez-Crespo 2025a, 4). In the context of AI, this includes the ability to adjust automation levels to task requirements and individual preferences (O'Brien 2024, 401). It also involves version control and role-aware permissions that ensure terminologists retain ultimate authority over terminological outputs and can override AI suggestions (Jiménez-Crespo 2025a, 6), thereby preventing other actors or AI systems from modifying data without explicit authorization from the terminologist (Warburton 2021, 174).

Preliminary evidence from Jiménez-Crespo's (2025a, 15) study on translators' preferences suggests that adaptive and interactive technologies may also strengthen terminologists' perceived autonomy and agency. His findings align with Briva-Iglesias (2024), who shows that, compared with traditional post-editing, interactive post-editing increases translators' sense of control while also improving productivity and fluency.

As for AI design for augmentation, a key concept already discussed in translation studies is cognitive ergonomics (Teixeira and O'Brien 2019). In HCI, cognitive ergonomics focuses on the mental processes involved in system use and seeks to reduce cognitive load and work stress while supporting decision-making and skilled performance (International Ergonomics Association 2018). TW is cognitively demanding; AI systems should, therefore, be designed to reduce unnecessary cognitive load and support terminologists' decision-making. These goals can be achieved through clear, well-structured interfaces and the automation of repetitive or mechanical tasks.

In AI-automated scenarios, decision-making support can be operationalized through explainability, i.e., a set of methods that help users understand AI outputs (Capel and Brereton 2023, 6). Although LLMs remain black boxes with opaque internal processes (H. Zhao et al. 2024, 2), partial workarounds exist. RAG, chain-of-thought prompting, and reasoning models can facilitate terminologists' assessment of the reliability of outputs.

RAG is arguably the most effective technique for justifying LLM outputs because it enables reference to verifiable online sources or user-supplied corpora. Given translators' interest in this functionality (Jiménez-Crespo 2025a, 16), it is likely to benefit terminologists as well. However, RAG is reliable only when it draws on credible sources. The optimal scenario is one in which terminologists supply a well-curated corpus. In contrast, if the terminologist cannot restrict the sources accessed by RAG, web-based implementations risk relying on inadequate content. This includes texts that are outdated or irrelevant to the intended domain (Heinisch 2025b, 70). RAG may also retrieve material selected not for its quality but due to optimization for generative engines, a strategy that increases retrieval likelihood (Aggarwal et al. 2024). As a result, LLM outputs may be grounded in unreliable sources, including AI-generated content.

While RAG is valuable, LLM suggestions should not rely on it exclusively, as this would prevent models from leveraging insights derived from their training data. LLMs can produce useful contributions not directly traceable to specific sources; in such cases, chain-of-thought prompting and reasoning models may provide an approximate justification.

Finally, the design of AI-enabled systems should address two additional automation risks: reduced critical engagement and anchoring. Reduced critical engagement denotes a decline in how rigorously users evaluate and validate AI outputs (Heer 2019, 1844; O'Brien 2024, 403). It may result from fatigue, time pressure, or an automation-induced illusion of reliability that discourages verification (Klingbeil et al. 2024, 7). Interface design can mitigate this risk by introducing intentional friction, such as confirmation dialogs that prompt terminologists to reassess AI suggestions before validation.

Closely related is the anchoring effect, whereby initial AI suggestions constrain subsequent human judgment and limit consideration of alternatives (Furnham and Boo 2011). This risk is especially pronounced in post-editing TW scenarios, where terminologists revise and validate AI suggestions. The primary countermeasure is to delay AI suggestions until after the terminologist has completed an initial independent analysis.

Taken together, these design options should be regarded as provisional hypotheses whose practical value depends on systematic empirical evaluation. Advancing AI-supported TW therefore hinges on sustained, participatory engagement with terminologists and on the rigorous application of established HCI methods, including focus groups, interviews, usability studies, and observational research (Capel and Brereton 2023, 4).

## 5. Conclusions

The integration of GenAI into TW is already underway, but how it is adopted is neither neutral nor predetermined. This paper has argued that AI can only be meaningfully integrated into TW if it is guided by a human-centered orientation that preserves professional agency and well-being. Terminologists must remain central actors in AI-assisted workflows, retaining authority over the management and dissemination of terms and the knowledge they convey. Without such an HCAI orientation, efficiency-driven deployments risk undermining the very expertise that ensures the reliability and adequacy of terminological products.

By presenting a human-centered approach, the paper has articulated a coherent framework organized around three dimensions: the augmented terminologist, ethical AI, and HCD. Together, these dimensions highlight that automation and human control are not opposing goals, that the creation of high-quality unbiased terminological resources requires the active involvement of terminologists, and that AI-enhanced tool and workflow design must be grounded in terminologists' needs.

Advancing a human-centered approach depends on a sustained empirical research agenda that examines how terminologists interact with AI in practice, determines which forms of automation genuinely augment the terminologist's capabilities, and systematically evaluates the terminological competence of LLMs under different conditions. This evidence is essential to ensuring that AI supports TW in ways that are compatible with the field's foundational values.

A human-centered orientation is essential to ensure that AI empowers, rather than displaces, terminologists. However, the implications of adopting AI extend beyond the profession itself, as they also affect the broader social value of TW. Terminological products shape specialized communication, translation, standardization, and the circulation of knowledge across languages and cultures. The decisions made today about how AI is integrated into TW will ultimately influence the preservation of accuracy and diversity in terminology and specialized knowledge.

## Funding

This research was funded by Canada's Social Sciences and Humanities Research Council, grant number 430-2023-0248, and Spain's Ministry of Science and Innovation, grant number PID2020-118369GB-I00.

## Acknowledgments

The author thanks Pamela Faber, Pilar León-Araúz, Catherine Trekker, and Gilles de Larminat for their comments on an earlier version of the article, as well as Kara Warburton for the suggestions she provided on the revised manuscript, after the peer-review process, in her capacity as editor of this special issue.

## Address for correspondence

Antonio San Martín
Université du Québec à Trois-Rivières
Canada

antonio.san.martin.pizarro@uqtr.ca

## Publication history